\documentclass[runningheads]{llncs}
\usepackage[T1]{fontenc}
\usepackage{graphicx}
\usepackage{cite}
\usepackage{booktabs}
\usepackage{multirow}
\usepackage{array}
\usepackage{xcolor}
\usepackage{amsfonts,amssymb,amsmath}
\usepackage{adjustbox}
\usepackage[colorlinks=true,linkcolor=black,citecolor=black,urlcolor=blue]{hyperref}
\usepackage{marvosym}
\usepackage{hyperref}

\begin{document}
\title{AdaSurvMamba: Dynamic Fusion and Semantic Scanning for Multimodal Survival Analysis}
\titlerunning{AdaSurvMamba}
%
\author{
  Jialong Zhong\inst{1}
  \index{Zhong, Jialong} 
  \and
  Tingwei Liu\inst{1}
  \index{Liu, Tingwei}
  \and
  Baokun Yue\inst{1}
  \index{Yue, Baokun} 
  \and 
  Jingjing Li\inst{2}
  \index{Li, Jingjing}
  \and
  Yongri Piao\inst{1}\textsuperscript{(\Letter)}
  \index{Piao, Yongri} 
  \and 
  Miao Zhang\inst{1}\textsuperscript{(\Letter)}
  \index{Zhang, Miao}  
  \and  
  Leiye Liu\inst{1}
  \index{Liu, Leiye} 
  \and
  Jiahong Jiang\inst{1}
  \index{Jiang, Jiahong} 
  \and  
  Wei Ji\inst{3}
  \index{Ji, Wei} 
  \and 
  Huchuan Lu\inst{1}
  \index{Lu, Huchuan} 
}

\authorrunning{J. Zhong et al.}
\institute{
  Dalian University of Technology
  \and
  University of Alberta
  \and
  Yale University \\
\email{yrpiao@dlut.edu.cn, miaozhang@dlut.edu.cn}}

\maketitle              
\begin{abstract}
Multimodal survival analysis utilizing whole slide images (WSIs) and genomic profiles is fundamental for cancer prognosis. Recently, state-space models like Mamba have emerged as powerful tools for sequence modeling. However, translating this success to complex multimodal tasks is hindered by two critical limitations. First, conventional fusion strategies assume a static multimodal interaction strength, ignoring the fluctuating diagnostic importance of each modality across different patients and local regions. Second, the standard Mamba architecture processes tokens along predefined physical paths. This rigid scanning disrupts the semantic continuity of spatially scattered medical features and exacerbates long-range decay. To address these challenges, we introduce AdaSurvMamba as a novel adaptive framework for multimodal survival analysis. The framework features a Dual-Scale Importance-Aware Reconstruction (DSIR) module to dynamically modulate cross-modal interaction strength. It evaluates diagnostic importance at both the sequence and token levels to reconstruct the input representations. Furthermore, we propose a Semantic Aggregation Scanning (SAS) module to overcome contextual fragmentation. The SAS module dynamically reorganizes discrete tokens into semantically continuous sequences via a shared prototype pool. It explicitly modulates the state transition step size using global modality context and semantic priors to adaptively control the information absorption rate. Experiments across five TCGA cohorts demonstrate consistent gains over existing methods. Code is available at \href{https://github.com/zjlGO/AdaSurvMamba}{https://github.com/zjlGO/AdaSurvMamba}.

\keywords{Survival outcome prediction  \and Whole slide image \and Mamba.}
\end{abstract}
 
\section{Introduction}
Deep learning has greatly advanced WSI analysis in computational pathology \cite{zhong2026diffusion, zheng2025gmmamba}. Survival analysis is a fundamental task in cancer prognosis and treatment planning \cite{song2023artificial}. Accurate prediction requires a comprehensive understanding of the tumor, which further motivates the use of multimodal data in clinical decision-making. Specifically, Whole Slide Images (WSIs) reveal detailed tumor morphology, while genomic profiles provide critical insights into molecular alterations. These two modalities offer distinct yet highly complementary prognostic clues \cite{song2024multimodal, wang2025histo}. Consequently, multimodal deep learning approaches that integrate WSIs and genomics have recently achieved remarkable success in survival prediction \cite{chen2021multimodal, ding2024multimodal, chen2022pan}.

Many early approaches in this field rely on late fusion to integrate these diverse features \cite{chen2020pathomic, chen2022pan}. While effective to some extent, these strategies often fail to capture critical cross-modal synergies. Recent methods have attempted to address this limitation by introducing attention mechanisms or hybrid supervision to better align pathological and genomic features \cite{chen2021multimodal, ding2024multimodal, fu2025hsfsurv, liu2025murrenet}. In parallel, the Mamba architecture has achieved notable success in sequence modeling tasks, such as whole slide image classification \cite{zhang20252dmamba, huang2024unleash, zheng2025m3amba, zheng2025graphmamba}. This success has inspired initial attempts to apply Mamba to multimodal survival analysis, leading to frameworks like SurvMamba \cite{chen2024survmamba} and ME-Mamba \cite{zhang2026me}. Unlike prior survival models using fixed cross-modal interactions or predefined scans, DSIR learns patient- and token-level modality relevance, while SAS performs semantic reordering before scanning. This also differs from visual Mamba variants that mainly redesign spatial scan paths for regular images, as our design targets sparse WSI regions and unordered genomic features.

Another challenge is the disruption of semantic continuity during the scanning process. Standard Mamba models process features along predefined scanning paths. However, the specific arrangement of input tokens profoundly impacts the sequence modeling performance \cite{huang2024localmamba, liu2025defmamba}. In WSIs, biologically related tissue patches often scatter across distant locations. Genomic features also lack a natural sequential order. Processing these inputs in a fixed order forces semantically connected elements to be far apart in the sequence. Since Mamba suffers from long-range decay \cite{gu2024mamba}, this forced distance prevents effective interactions and causes severe semantic fragmentation. Consequently, how to gather these discrete features into semantically connected sequences remains a critical problem to address.

To address these challenges, we propose AdaSurvMamba, a novel adaptive framework for multimodal survival analysis. It dynamically controls feature interactions and maintains semantic continuity. The framework contains a Dual-Scale Importance-Aware Reconstruction (DSIR) module and a Semantic Aggregation Scanning (SAS) module. The DSIR module overcomes the rigidity of existing fusion methods. It evaluates diagnostic importance at both the sequence and token levels to rebuild the input representations. This reconstruction directly regulates the multimodal interaction strength. It allows the model to focus on the most critical prognostic signals. Building upon this, the SAS module overcomes the issue of semantic fragmentation. It further reorganizes the rebuilt features into semantically continuous scanning sequences. This keeps related tokens connected during the state transition process and avoids the drawbacks of predefined scanning paths. Together, the DSIR and SAS modules enable AdaSurvMamba to flexibly integrate features and capture robust context. Extensive experiments on multiple public datasets validate the effectiveness of our framework.

\section{Method}
\begin{figure}[t!]
\centering
\includegraphics[width=\textwidth,keepaspectratio]{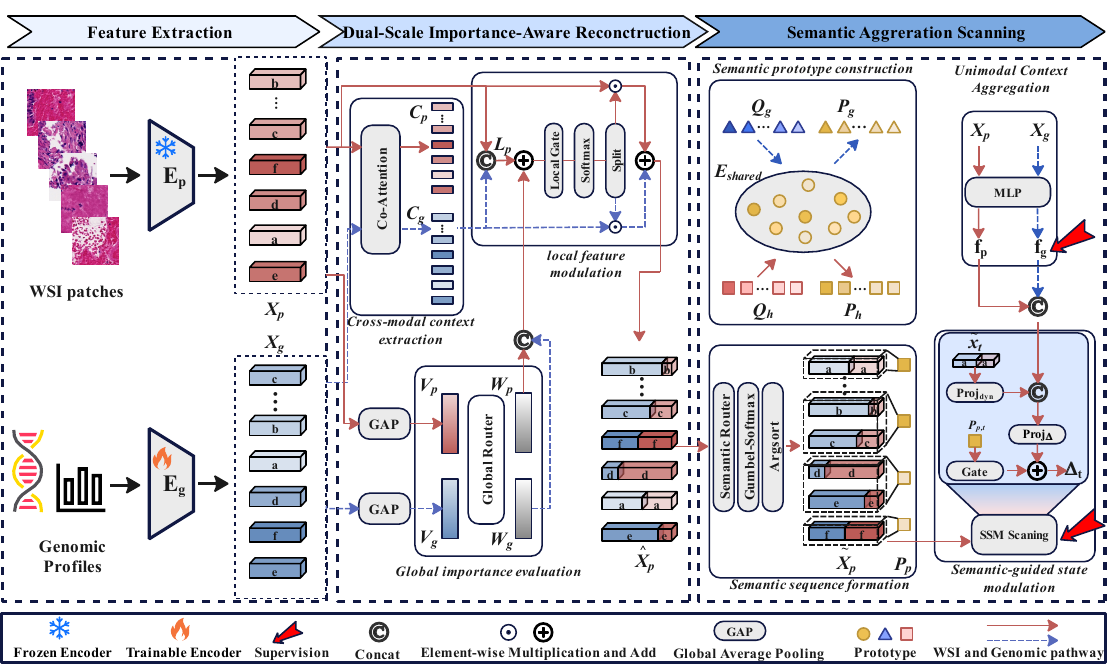} 
\caption{Overview of the proposed AdaSurvMamba framework. For visual clarity, the detailed computational flow within the DSIR and SAS modules is primarily illustrated using the pathology (WSI) branch, while the genomic branch follows a symmetric processing paradigm.}
\label{fig1}
\end{figure}

Figure~\ref{fig1} illustrates the overall architecture of AdaSurvMamba. The framework processes WSI patches and genomic profiles. A frozen foundation model encodes WSI patches, while a learnable SNN maps six functional groups derived from RNA-seq, CNV, and mutation data into six tokens $\mathbf{X}_g\in\mathbb{R}^{6\times D}$. The Dual-Scale Importance-Aware Reconstruction (DSIR) module then computes global and local importance weights to dynamically adjust these multimodal features. Next, the Semantic Aggregation Scanning (SAS) module reorganizes the adjusted tokens into semantically coherent sequences for state space modeling. Finally, the network aggregates the enhanced representations to predict patient survival.

\subsection{Preliminaries}
Mamba formulates sequence modeling through structured state space models (SSMs). The continuous dynamics map an input $x(t)$ to output $y(t)$ via a hidden state $h(t)$:
\begin{equation}
h'(t) = \mathbf{A}h(t) + \mathbf{B}x(t), \quad y(t) = \mathbf{C}h(t) + \mathbf{D}x(t),
\end{equation}
where $\mathbf{A}$, $\mathbf{B}$, $\mathbf{C}$, and $\mathbf{D}$ are learnable matrices. Discretization converts this process into a recurrent form:
\begin{equation}
h_t = \mathbf{\bar{A}} h_{t-1} + \mathbf{\bar{B}} x_t.
\end{equation}
The key innovation lies in its selectivity mechanism. The parameters $\mathbf{B}$, $\mathbf{C}$, and the timescale $\Delta$ become input-dependent functions. This allows the model to selectively retain or forget information during state transitions. However, this recurrent process strictly follows the predefined physical order of input tokens. Such rigidity prevents effective modeling of long-range dependencies in multimodal medical data.

\subsection{Dual-Scale Importance-Aware Reconstruction}
Existing fusion methods apply static interaction weights across all samples. This overlooks the fact that modality importance varies both across patients and within local feature pairs. The DSIR module addresses this by dynamically reconstructing features at two complementary scales.

\textbf{Cross-modal context extraction.} We establish early interactions between pathological tokens $\mathbf{X}_p \in \mathbb{R}^{N \times D}$ and genomic tokens $\mathbf{X}_g \in \mathbb{R}^{M \times D}$. To avoid the heavy computational burden of standard cross-attention, we employ an efficient co-attention mechanism. It calculates a joint affinity matrix from the linearly projected modalities to derive attention vectors, which then explicitly modulate the original tokens. This produces context-aware features $\mathbf{C}_p \in \mathbb{R}^{N \times D}$ and $\mathbf{C}_g \in \mathbb{R}^{M \times D}$, effectively bridging the initial semantic gap with minimal overhead.

\textbf{Global importance evaluation.} The dominant prognostic modality often differs across patients. We quantify this macroscopic preference through a global confidence router. Global average pooling condenses $\mathbf{X}_p$ and $\mathbf{X}_g$ into vectors $\mathbf{v}_p, \mathbf{v}_g \in \mathbb{R}^D$. A multi-layer perceptron processes their concatenation to produce sequence-level importance weights $\mathbf{w}_p, \mathbf{w}_g \in \mathbb{R}^D$. These weights reflect the overall contribution of each modality for the current patient.

\textbf{Local feature modulation.} Even after global adjustment, the importance of individual token pairs still varies. We model this microscopic variation through a local fusion network. Taking the pathological branch as an example, the network processes the concatenated tensor $[\mathbf{X}_p, \mathbf{C}_g]$ to produce initial gate logits $\mathbf{L}_p \in \mathbb{R}^{N \times 2 \times D}$. We then inject the global importance weights to obtain the final gating signals:
\begin{equation}
\tilde{\mathbf{L}}_p[:, 0, :] = \mathbf{L}_p[:, 0, :] + \mathbf{w}_p, \quad \tilde{\mathbf{L}}_p[:, 1, :] = \mathbf{L}_p[:, 1, :] + \mathbf{w}_g.
\end{equation}
A softmax operation applied along the modality dimension yields the dynamic gates $\mathbf{G}_{raw}$ and $\mathbf{G}_{ctx}$. The reconstructed pathological representation becomes:
\begin{equation}
\hat{\mathbf{X}}_p = \mathbf{X}_p \odot \mathbf{G}_{raw} + \mathbf{C}_g \odot \mathbf{G}_{ctx}.
\end{equation}
The genomic branch undergoes a symmetric process to produce $\hat{\mathbf{X}}_g \in \mathbb{R}^{M \times D}$. This dual-scale strategy amplifies critical signals before they enter the SSM.

\subsection{Semantic Aggregation Scanning}
Standard Mamba processes tokens along fixed physical paths. This disrupts the semantic continuity of medical features that are spatially scattered in the original data. The SAS module solves this by dynamically reorganizing tokens and modulating state transitions based on semantic content.

\textbf{Semantic prototype construction.} Pathology and genomics have distinct information structures. We build a shared prototype pool $\mathbf{E}_{shared} \in \mathbb{R}^{r \times d_p}$ to map both modalities into a unified semantic space. Here $r$ is the rank for semantic decoupling and $d_p$ is the prototype dimension. For modality $m \in \{p,g\}$, we learn a coefficient matrix $\mathbf{Q}_m \in \mathbb{R}^{T_m \times r}$ to extract modality-specific prototypes:
\begin{equation}
\mathbf{P}_m = \mathbf{Q}_m \mathbf{E}_{shared} \in \mathbb{R}^{T_m \times d_p},
\end{equation}
where $T_m$ is the number of prototypes for that modality. This design captures modal specifics while enforcing cross-modal alignment through the shared pool. Since genomics is compressed into six pathway-level tokens, fewer prototypes summarize major molecular programs without over-fragmentation, whereas numerous heterogeneous WSI patches require finer grouping. This is an ablation-supported design choice rather than an assumption of limited genomic heterogeneity.

\textbf{Semantic sequence formation.} A linear router and Gumbel-Softmax generate discrete routing weights $\mathbf{R}_m \in \mathbb{R}^{L_m \times T_m}$ for each token in $\hat{\mathbf{X}}_m$. These weights serve a dual architectural purpose. Spatially, they dictate an argsort operation to deterministically cluster identically assigned tokens, forming the continuous sequence $\tilde{\mathbf{X}}_m$. Semantically, they are multiplied with the prototype pool to extract a differentiable context bias. By explicitly injecting this bias into the continuous SSM step size ($\Delta_t$), the model establishes a valid gradient pathway back to the router, thereby optimizing the discrete semantic reorganization end-to-end.

\textbf{Unimodal Context Aggregation.} Before entering the SAS module, each modality branch independently aggregates its reconstructed sequence via attention pooling. This produces global feature vectors $\mathbf{f}_p \in \mathbb{R}^D$ and $\mathbf{f}_g \in \mathbb{R}^D$, which capture the overall diagnostic signal of pathology and genomics for the current patient. These vectors are also passed through a classifier to generate auxiliary hazard predictions $\hat{\mathbf{h}}_p$ and $\hat{\mathbf{h}}_g$, providing additional supervision. The global features serve as compact representations of modality-level importance.

\textbf{Semantic-guided state modulation.} In state space models, the step size $\Delta$ governs how much new information flows into the hidden state. We explicitly guide this flow using both the semantic prototypes and the global modality context. First, we compress each token feature $\hat{\mathbf{x}}_t$ (the $t$-th row of $\tilde{\mathbf{X}}_m$) into a low-dimensional dynamic vector $\mathbf{x}_t^{\mathrm{dyn}}$ via a linear projection $\mathrm{Proj}_{\mathrm{dyn}}$:
\begin{equation}
\mathbf{x}_t^{\mathrm{dyn}} = \mathrm{Proj}_{\mathrm{dyn}}(\hat{\mathbf{x}}_t) \in \mathbb{R}^{d_{\mathrm{dyn}}}.
\end{equation}
We then concatenate $\mathbf{x}_t^{\mathrm{dyn}}$ with the global features $\mathbf{f} = [\mathbf{f}_p; \mathbf{f}_g] \in \mathbb{R}^{2D}$ and feed the result into a step size generator $\mathrm{Proj}_{\Delta}$, a linear layer that outputs the base step size:
\begin{equation}
\Delta_{\mathrm{base}, t} = \mathrm{Proj}_{\Delta}\big( [\mathbf{x}_t^{\mathrm{dyn}}; \mathbf{f}] \big) \in \mathbb{R}^{d_{\Delta}}.
\end{equation}
This injects patient-level modality dominance into the token-wise state transition. Meanwhile, we compute a prototype bias by aggregating the modality-specific prototypes according to the routing weights:
\begin{equation}
\mathbf{B}_{\mathrm{proto}, t} = \mathrm{Gate}\big( \mathbf{R}_{m,t} \mathbf{P}_m \big) \in \mathbb{R}^{d_{\Delta}},
\end{equation}
where $\mathbf{R}_{m,t} \in \mathbb{R}^{T_m}$ is the routing vector for token $t$, and $\mathrm{Gate}$ is a linear layer followed by $\tanh$ activation. The final semantic-guided step size is the sum of both contributions:
\begin{equation}
\Delta_t = \Delta_{\mathrm{base}, t} + \mathbf{B}_{\mathrm{proto}, t}.
\end{equation}
We apply the selective scan using $\Delta_t$ on the reordered sequence $\tilde{\mathbf{X}}_m$. This ensures that tokens belonging to the same semantic group receive coherent state updates, and that the update intensity is further modulated by the overall importance of each modality. Following the scan, the enhanced token representations are aggregated into a unified patient profile and fed into a prediction head to estimate the discrete-time survival hazard probabilities.

\textbf{Optimization objective.} Following \cite{chen2021multimodal}, we apply discrete-time negative log-likelihood to the fused and two unimodal predictions. The total objective is $\mathcal{L}=\mathcal{L}_{\mathrm{main}}+\lambda(\mathcal{L}_{\mathrm{aux}}^{p}+\mathcal{L}_{\mathrm{aux}}^{g})$, where the auxiliary terms preserve modality-specific prognostic information.

\section{Experiments}
\subsection{Experiment Protocol}
\textbf{Datasets.} We evaluated AdaSurvMamba on five TCGA \cite{kandoth2013mutational} benchmark cohorts: BLCA ($n=373$), BRCA ($n=956$), GBMLGG ($n=569$), LUAD ($n=453$), and UCEC ($n=480$). Each patient comprises diagnostic WSIs and paired genomic profiles (RNA-Seq, copy number variation, and somatic mutations). Following established protocols, we grouped the genomic features into six functional categories.

\textbf{Implementation details.} WSIs were cropped into $512 \times 512$ patches at $20\times$ magnification. Pre-trained foundation models (ResNet-50 \cite{he2016deep} and UNI \cite{chen2024towards}) extracted the instance-level morphological features. For the proposed SAS module, the semantic decoupling rank $r$ and the prototype dimension $d_p$ were set to 32 and 64, respectively. Based on our ablation studies, the optimal number of prototypes for genomics and pathology were configured as $T_g=4$ and $T_p=16$. We implemented the framework in PyTorch on a single RTX 4090 GPU. The network was optimized using Adam with a learning rate of $2 \times 10^{-4}$ and weight decay of $1 \times 10^{-5}$. The auxiliary loss weight $\lambda$ was empirically set to 0.5 to balance fused survival prediction and modality-specific supervision. We report the mean Concordance Index (C-Index) and standard deviation from 5-fold cross-validation.

\begin{table}[t!]
\centering
\scriptsize
  \caption{Results of different methods. The best results are shown in bold, and the second-best results are underlined.}
  \label{tab1}
\begin{adjustbox}{max width=\linewidth}
\begin{tabular}{lcccccc}
\toprule
\textbf{Method} &\textbf{Modality}  & \textbf{BLCA} & \textbf{BRCA} & \textbf{LUAD} & \textbf{GBMLGG} & \textbf{UCEC}

  \\ \midrule[1pt] \midrule[1pt]
  &\multicolumn{6}{c}{\textbf{Genomic Profiles}}
 \\ \midrule[0.5pt]
MLP & Genomics & 0.611$\pm$0.030 & 0.619$\pm$0.053 & 0.619$\pm$0.044 & 0.779$\pm$0.035 & 0.635$\pm$0.069\\
SNN \cite{klambauer2017self} & Genomics & 0.633$\pm$0.074 & 0.634$\pm$0.051 & 0.622$\pm$0.041 & 0.816$\pm$0.028 & 0.620$\pm$0.039
  
  \\ \midrule[1pt] \midrule[1pt]
  &\multicolumn{6}{c}{\textbf{Feature extractor: ResNet50 \cite{he2016deep}}}
 \\ \midrule[0.5pt]
ABMIL \cite{ilse2018attention} & WSI & 0.589$\pm$0.058 & 0.577$\pm$0.084 & 0.567$\pm$0.042 & 0.771$\pm$0.033 & 0.668$\pm$0.019\\
Transmil \cite{shao2021transmil} & WSI & 0.579$\pm$0.046 & 0.640$\pm$0.035 & 0.585$\pm$0.039 & 0.705$\pm$0.094 & 0.597$\pm$0.086
  \\ \midrule[0.5pt]

MCAT \cite{chen2021multimodal} & Multimodal & \underline{0.677$\pm$0.036} & 0.657$\pm$0.048 & 0.662$\pm$0.035 &\underline{0.847$\pm$0.042} &0.675$\pm$0.045\\
MOTCAT \cite{xu2023multimodal} & Multimodal & 0.677$\pm$0.024 & 0.677$\pm$0.020 & \underline{0.678$\pm$0.033} &0.841$\pm$0.015 &0.672$\pm$0.042\\
MMP \cite{song2024multimodal} & Multimodal & 0.626$\pm$0.042 & 0.608$\pm$0.035 & 0.650$\pm$0.049 & 0.816$\pm$0.021 & 0.648$\pm$0.069\\
CCL \cite{zhou2024cohort} & Multimodal & 0.652$\pm$0.038 & 0.596$\pm$0.054 & 0.640$\pm$0.057 & 0.845$\pm$0.014 & 0.686$\pm$0.046\\
GHANT \cite{wang2025histo} & Multimodal & 0.605$\pm$0.016 & 0.674$\pm$0.071 & 0.595$\pm$0.059 & 0.800$\pm$0.024 & 0.652$\pm$0.039\\
LD-VAE \cite{zhou2025robust} & Multimodal & 0.614$\pm$0.026 & 0.632$\pm$0.065 & 0.638$\pm$0.035 & 0.841$\pm$0.026 & \underline{0.694$\pm$0.046}\\
MGCM \cite{cui2026mgcm} & Multimodal & 0.675$\pm$0.024 & \underline{0.681$\pm$0.057} & 0.669$\pm$0.059 & 0.845$\pm$0.024 & 0.688$\pm$0.039\\
Ours & Multimodal & \textbf{0.707$\pm$0.029} & \textbf{0.698$\pm$0.049} & \textbf{0.691$\pm$0.040} & \textbf{0.859$\pm$0.023} & \textbf{0.707$\pm$0.046} 

\\ \midrule[1pt] \midrule[1pt]
  &\multicolumn{6}{c}{\textbf{Feature extractor: UNI \cite{chen2024towards}}}
  \\ \midrule[0.5pt]
ABMIL \cite{ilse2018attention} & WSI & 0.611$\pm$0.034 & 0.630$\pm$0.043 & 0.640$\pm$0.016 & 0.803$\pm$0.047 & 0.644$\pm$0.066\\
Transmil \cite{shao2021transmil} & WSI & 0.590$\pm$0.073 & 0.642$\pm$0.037 & 0.561$\pm$0.066 & 0.735$\pm$0.048 & 0.634$\pm$0.064
  \\ \midrule[0.5pt]
MCAT \cite{chen2021multimodal} & Multimodal & 0.682$\pm$0.052 & 0.667$\pm$0.023 & 0.671$\pm$0.036 & 0.842$\pm$0.016 & 0.688$\pm$0.041\\
MOTCAT \cite{xu2023multimodal} & Multimodal & 0.678$\pm$0.026 & \underline{0.690$\pm$0.019} & 0.673$\pm$0.028 & 0.840$\pm$0.023 & 0.690$\pm$0.050\\
MMP \cite{song2024multimodal} & Multimodal & 0.611$\pm$0.029 & 0.592$\pm$0.016 & 0.648$\pm$0.035 & 0.815$\pm$0.023 &0.654$\pm$0.056\\
CCL \cite{zhou2024cohort} & Multimodal & 0.657$\pm$0.034 & 0.620$\pm$0.051 & 0.649$\pm$0.056 & \underline{0.852$\pm$0.032} & 0.682$\pm$0.050\\
GHANT \cite{wang2025histo} & Multimodal & 0.622$\pm$0.042 & 0.667$\pm$0.029 & 0.608$\pm$0.059 & 0.808$\pm$0.016 & \underline{0.705$\pm$0.036}\\
LD-VAE \cite{zhou2025robust} & Multimodal & 0.650$\pm$0.015 & 0.673$\pm$0.026 & 0.643$\pm$0.057 & 0.833$\pm$0.025 & 0.696$\pm$0.053\\
MGCM \cite{cui2026mgcm} & Multimodal & \underline{0.695$\pm$0.024} & 0.689$\pm$0.027 & \underline{0.679$\pm$0.033} & 0.850$\pm$0.021 & 0.702$\pm$0.039\\
Ours & Multimodal & \textbf{0.723$\pm$0.026} & \textbf{0.701$\pm$0.039} & \textbf{0.696$\pm$0.032} & \textbf{0.866$\pm$0.017} & \textbf{0.718$\pm$0.038} 

\\ \midrule[1pt]
\bottomrule
\end{tabular}
\end{adjustbox}
\end{table}

\subsection{Comparison with State-of-the-Arts}
Table~\ref{tab1} presents the performance comparison of AdaSurvMamba against existing methods across five TCGA cohorts. We include both unimodal baselines (using only genomics or only WSIs) and state-of-the-art multimodal fusion approaches. For genomic-only methods, we compare with MLP and SNN \cite{klambauer2017self}. For WSI-only methods, we consider ABMIL \cite{ilse2018attention} and Transmil \cite{shao2021transmil}. For multimodal methods, we evaluate against advanced frameworks including MCAT \cite{chen2021multimodal}, MOTCAT \cite{xu2023multimodal}, MMP \cite{song2024multimodal}, CCL \cite{zhou2024cohort}, GHANT \cite{wang2025histo}, LD-VAE \cite{zhou2025robust}, and MGCM \cite{cui2026mgcm}. SurvMamba \cite{chen2024survmamba} and ME-Mamba \cite{zhang2026me} are not included because their official implementations were not publicly available at submission time, preventing reproduction under our unified protocol. As shown in Table~\ref{tab1}, AdaSurvMamba achieves consistent improvements over these competitors across all datasets. Specifically, under the UNI setting, our framework achieves an average C-Index improvement of 2.32\% over the best competing method across all five cohorts. Similar gains are observed with ResNet50 features, where the average improvement reaches 2.43\%. These consistent performance gains highlight the effectiveness of our adaptive multimodal interaction strategy for accurate survival prediction.

\subsection{Ablation Study}
We conduct extensive ablation studies on the BLCA and BRCA cohorts using ResNet50 features to systematically validate our proposed designs. The baseline model is a standard Mamba architecture utilizing straightforward concatenation fusion and default 1D physical scanning.

\textbf{Effectiveness of Core Modules.} Table~\ref{tab:ablation_modules} isolates the macroscopic contributions of our architectural components. The baseline yields suboptimal performance due to its static fusion and rigid scanning paths. Integrating the DSIR module notably improves the C-Index, demonstrating the necessity of dynamically modulating multimodal interactions. Furthermore, employing the SAS module effectively mitigates long-range decay by reorganizing scattered features into semantically continuous sequences, achieving the best overall performance when combined.

\textbf{Internal Mechanisms of DSIR.} We investigate the fine-grained design of the DSIR module in Table~\ref{tab:ablation_mod_strategy}. We compare our global-guided dynamic gating against naive modulation strategies. Our gating mechanism significantly outperforms simple scalar shifting or additive attention. This indicates that a simple spatial shift is insufficient to model microscopic token-pair relevance, whereas our softmax-based gating decisively amplifies critical local prognostic signals under macroscopic guidance.

\textbf{Internal Mechanisms of SAS.} We deeply analyze the SAS module from two perspectives: scanning strategy and semantic capacity. First, Table~\ref{tab:ablation_scanning} compares our semantic scanning against alternative scanning paths, including conventional 1-way (sequential) and 2-way (bidirectional) physical scans, as well as a random shuffle scan. While 2-way scanning marginally improves upon the 1-way baseline by capturing bidirectional physical contexts, it still suffers from spatial rigidity. Conversely, the random shuffle strategy severely disrupts inherent local continuity, leading to performance degradation. Our semantic scanning, however, leverages learned prototypes to meaningfully reorganize tokens. It significantly outperforms all physical and random scanning paths by constructing semantically coherent sequences, while requiring only a single efficient forward pass. Second, Table~\ref{tab:ablation_prototypes} analyzes the impact of modality-specific prototype sizes ($T_g, T_p$). The best setting ($T_g=4,T_p=16$) reflects the different token scales rather than limited genomic heterogeneity.

\begin{table}[htbp]
\centering
\scriptsize
\begin{minipage}[t]{0.48\linewidth}
  \centering
  \caption{Ablation of core modules.}
  \label{tab:ablation_modules}
  \begin{adjustbox}{max width=\linewidth}
  \begin{tabular}{cc|cc}
  \toprule
  \textbf{DSIR} & \textbf{SAS} & \textbf{BLCA} & \textbf{BRCA} \\ 
  \midrule[1pt]
  - & - & 0.642$\pm$0.035 & 0.645$\pm$0.041 \\
  $\checkmark$ & - & 0.675$\pm$0.032 & 0.668$\pm$0.045 \\
  - & $\checkmark$ & 0.669$\pm$0.028 & 0.661$\pm$0.038 \\
  $\checkmark$ & $\checkmark$ & \textbf{0.707$\pm$0.029} & \textbf{0.698$\pm$0.049} \\
  \bottomrule
  \end{tabular}
  \end{adjustbox}
\end{minipage}\hfill
\begin{minipage}[t]{0.48\linewidth}
  \centering
  \caption{DSIR modulation strategies.}
  \label{tab:ablation_mod_strategy}
  \begin{adjustbox}{max width=\linewidth}
  \begin{tabular}{l|cc}
  \toprule
  \textbf{Strategy} & \textbf{BLCA} & \textbf{BRCA} \\ 
  \midrule[1pt]
  Concat (No DSIR) & 0.669$\pm$0.028 & 0.661$\pm$0.038 \\
  Additive Attn & 0.681$\pm$0.034 & 0.674$\pm$0.041 \\
  Simple Scalar & 0.693$\pm$0.031 & 0.685$\pm$0.045 \\
  DSIR Gating & \textbf{0.707$\pm$0.029} & \textbf{0.698$\pm$0.049} \\
  \bottomrule
  \end{tabular}
  \end{adjustbox}
\end{minipage}

\vspace{4mm} 

\begin{minipage}[t]{0.48\linewidth}
  \centering
  \caption{SAS scanning paths.}
  \label{tab:ablation_scanning}
  \begin{adjustbox}{max width=\linewidth}
  \begin{tabular}{l|cc}
  \toprule
  \textbf{Scanning Path} & \textbf{BLCA} & \textbf{BRCA} \\ 
  \midrule[1pt]
  1-way Physical & 0.675$\pm$0.032 & 0.668$\pm$0.045 \\
  2-way Physical & 0.684$\pm$0.030 & 0.679$\pm$0.042 \\
  Random Shuffle & 0.661$\pm$0.036 & 0.655$\pm$0.050 \\
  Semantic Scan (Ours) & \textbf{0.707$\pm$0.029} & \textbf{0.698$\pm$0.049} \\
  \bottomrule
  \end{tabular}
  \end{adjustbox}
\end{minipage}\hfill
\begin{minipage}[t]{0.48\linewidth}
  \centering
  \caption{SAS prototype sizes $(T_g, T_p)$.}
  \label{tab:ablation_prototypes}
  \begin{adjustbox}{max width=\linewidth}
  \begin{tabular}{c|cc}
  \toprule
  \textbf{Sizes ($T_g, T_p$)} & \textbf{BLCA} & \textbf{BRCA} \\ 
  \midrule[1pt]
  (4, 4) & 0.688$\pm$0.031 & 0.681$\pm$0.044 \\
  (4, 8) & 0.697$\pm$0.030 & 0.692$\pm$0.045 \\
  (4, 16)& \textbf{0.707$\pm$0.029} & \textbf{0.698$\pm$0.049} \\
  (8, 8) & 0.692$\pm$0.028 & 0.686$\pm$0.045 \\
  \bottomrule
  \end{tabular}
  \end{adjustbox}
\end{minipage}
\end{table}

\section{Conclusion}
We presented AdaSurvMamba, an adaptive state-space framework for multimodal cancer survival prediction. By introducing DSIR for dynamic interaction modulation and SAS for semantic scanning with adaptive step-size control, our approach overcomes the limitations of static fusion and rigid physical scanning. Results on five TCGA cohorts show consistent gains. External validation remains future work because cohorts with matched WSIs, genomic profiles, and survival outcomes are still scarce.

\begin{credits} 
\subsubsection{\discintname}
We declared no competing interests.
\end{credits}

\bibliographystyle{splncs04}
\bibliography{paper-4611}

@article{song2023artificial,
  title={Artificial intelligence for digital and computational pathology},
  author={Song, Andrew H and Jaume, Guillaume and Williamson, Drew FK and Lu, Ming Y and Vaidya, Anurag and Miller, Tiffany R and Mahmood, Faisal},
  journal={Nature Reviews Bioengineering},
  volume={1},
  number={12},
  pages={930--949},
  year={2023},
  publisher={Nature Publishing Group UK London}
}

@article{song2024multimodal,
  title={Multimodal prototyping for cancer survival prediction},
  author={Song, Andrew H and Chen, Richard J and Jaume, Guillaume and Vaidya, Anurag J and Baras, Alexander S and Mahmood, Faisal},
  journal={arXiv preprint arXiv:2407.00224},
  year={2024}
}

@article{wang2025histo,
  title={Histo-genomic knowledge association for cancer prognosis from histopathology whole slide images},
  author={Wang, Zhikang and Zhang, Yumeng and Xu, Yingxue and Imoto, Seiya and Chen, Hao and Song, Jiangning},
  journal={IEEE Transactions on Medical Imaging},
  volume={44},
  number={5},
  pages={2170--2181},
  year={2025},
  publisher={IEEE}
}

@article{chen2020pathomic,
  title={Pathomic fusion: an integrated framework for fusing histopathology and genomic features for cancer diagnosis and prognosis},
  author={Chen, Richard J and Lu, Ming Y and Wang, Jingwen and Williamson, Drew FK and Rodig, Scott J and Lindeman, Neal I and Mahmood, Faisal},
  journal={IEEE transactions on medical imaging},
  volume={41},
  number={4},
  pages={757--770},
  year={2020},
  publisher={IEEE}
}

@article{chen2022pan,
  title={Pan-cancer integrative histology-genomic analysis via multimodal deep learning},
  author={Chen, Richard J and Lu, Ming Y and Williamson, Drew FK and Chen, Tiffany Y and Lipkova, Jana and Noor, Zahra and Shaban, Muhammad and Shady, Maha and Williams, Mane and Joo, Bumjin and others},
  journal={Cancer cell},
  volume={40},
  number={8},
  pages={865--878},
  year={2022},
  publisher={Elsevier}
}

@inproceedings{zhang20252dmamba,
  title={2dmamba: Efficient state space model for image representation with applications on giga-pixel whole slide image classification},
  author={Zhang, Jingwei and Nguyen, Anh Tien and Han, Xi and Trinh, Vincent Quoc-Huy and Qin, Hong and Samaras, Dimitris and Hosseini, Mahdi S},
  booktitle={Proceedings of the Computer Vision and Pattern Recognition Conference},
  pages={3583--3592},
  year={2025}
}

@article{huang2024unleash,
  title={Unleash the power of state space model for whole slide image with local aware scanning and importance resampling},
  author={Huang, Yanyan and Zhao, Weiqin and Fu, Yu and Zhu, Lingting and Yu, Lequan},
  journal={IEEE Transactions on Medical Imaging},
  volume={44},
  number={2},
  pages={1032--1042},
  year={2024},
  publisher={IEEE}
}

@article{chen2024survmamba,
  title={Survmamba: state space model with multi-grained multi-modal interaction for survival prediction},
  author={Chen, Ying and Xie, Jiajing and Lin, Yuxiang and Song, Yuhang and Yang, Wenxian and Yu, Rongshan},
  journal={arXiv preprint arXiv:2404.08027},
  year={2024}
}

@article{zhang2026me,
  title={ME-Mamba: Multi-Expert Mamba with efficient knowledge capture and fusion for multimodal survival analysis},
  author={Zhang, Chengsheng and Qu, Linhao and Liu, Xiaoyu and Song, Zhijian},
  journal={Computerized Medical Imaging and Graphics},
  volume={129},
  pages={102733},
  year={2026},
  publisher={Elsevier}
}

@inproceedings{gu2024mamba,
  title={Mamba: Linear-time sequence modeling with selective state spaces},
  author={Gu, Albert and Dao, Tri},
  booktitle={First conference on language modeling},
  year={2024}
}

@inproceedings{huang2024localmamba,
  title={Localmamba: Visual state space model with windowed selective scan},
  author={Huang, Tao and Pei, Xiaohuan and You, Shan and Wang, Fei and Qian, Chen and Xu, Chang},
  booktitle={European conference on computer vision},
  pages={12--22},
  year={2024},
  organization={Springer}
}

@inproceedings{liu2025defmamba,
  title={Defmamba: Deformable visual state space model},
  author={Liu, Leiye and Zhang, Miao and Yin, Jihao and Liu, Tingwei and Ji, Wei and Piao, Yongri and Lu, Huchuan},
  booktitle={Proceedings of the Computer Vision and Pattern Recognition Conference},
  pages={8838--8847},
  year={2025}
}

@inproceedings{chen2021multimodal,
  title={Multimodal co-attention transformer for survival prediction in gigapixel whole slide images},
  author={Chen, Richard J and Lu, Ming Y and Weng, Wei-Hung and Chen, Tiffany Y and Williamson, Drew FK and Manz, Trevor and Shady, Maha and Mahmood, Faisal},
  booktitle={Proceedings of the IEEE/CVF international conference on computer vision},
  pages={4015--4025},
  year={2021}
}

@article{ding2024multimodal,
  title={Multimodal co-attention fusion network with online data augmentation for cancer subtype classification},
  author={Ding, Saisai and Li, Juncheng and Wang, Jun and Ying, Shihui and Shi, Jun},
  journal={IEEE Transactions on Medical Imaging},
  volume={43},
  number={11},
  pages={3977--3989},
  year={2024},
  publisher={IEEE}
}

@article{kandoth2013mutational,
  title={Mutational landscape and significance across 12 major cancer types},
  author={Kandoth, Cyriac and McLellan, Michael D and Vandin, Fabio and Ye, Kai and Niu, Beifang and Lu, Charles and Xie, Mingchao and Zhang, Qunyuan and McMichael, Joshua F and Wyczalkowski, Matthew A and others},
  journal={Nature},
  volume={502},
  number={7471},
  pages={333--339},
  year={2013},
  publisher={Nature Publishing Group UK London}
}

@article{fu2025hsfsurv,
  title={HSFSurv: A hybrid supervision framework at individual and feature levels for multimodal cancer survival analysis},
  author={Fu, Bangkang and He, Junjie and Zhang, Xiaoli and Peng, Yunsong and Zhang, Zhuxu and Tang, Qi and Liu, Xinfeng and Cao, Ying and Wang, Rongpin},
  journal={Medical Image Analysis},
  pages={103810},
  year={2025},
  publisher={Elsevier}
}

@inproceedings{liu2025murrenet,
  title={Murrenet: Modeling holistic multimodal interactions between histopathology and genomic profiles for survival prediction},
  author={Liu, Mingxin and Cai, Chengfei and Li, Jun and Xu, Pengbo and Li, Jinze and Ma, Jiquan and Xu, Jun},
  booktitle={International Conference on Medical Image Computing and Computer-Assisted Intervention},
  pages={396--406},
  year={2025},
  organization={Springer}
}

@inproceedings{he2016deep,
  title={Deep residual learning for image recognition},
  author={He, Kaiming and Zhang, Xiangyu and Ren, Shaoqing and Sun, Jian},
  booktitle={Proceedings of the IEEE conference on computer vision and pattern recognition},
  pages={770--778},
  year={2016}
}

@article{chen2024towards,
  title={Towards a general-purpose foundation model for computational pathology},
  author={Chen, Richard J and Ding, Tong and Lu, Ming Y and Williamson, Drew FK and Jaume, Guillaume and Song, Andrew H and Chen, Bowen and Zhang, Andrew and Shao, Daniel and Shaban, Muhammad and others},
  journal={Nature medicine},
  volume={30},
  number={3},
  pages={850--862},
  year={2024},
  publisher={Nature Publishing Group US New York}
}

@inproceedings{ilse2018attention,
  title={Attention-based deep multiple instance learning},
  author={Ilse, Maximilian and Tomczak, Jakub and Welling, Max},
  booktitle={International conference on machine learning},
  pages={2127--2136},
  year={2018},
  organization={PMLR}
}

@article{shao2021transmil,
  title={Transmil: Transformer based correlated multiple instance learning for whole slide image classification},
  author={Shao, Zhuchen and Bian, Hao and Chen, Yang and Wang, Yifeng and Zhang, Jian and Ji, Xiangyang and others},
  journal={Advances in neural information processing systems},
  volume={34},
  pages={2136--2147},
  year={2021}
}

@inproceedings{xu2023multimodal,
  title={Multimodal optimal transport-based co-attention transformer with global structure consistency for survival prediction},
  author={Xu, Yingxue and Chen, Hao},
  booktitle={Proceedings of the IEEE/CVF international conference on computer vision},
  pages={21241--21251},
  year={2023}
}

@article{zhou2024cohort,
  title={Cohort-individual cooperative learning for multimodal cancer survival analysis},
  author={Zhou, Huajun and Zhou, Fengtao and Chen, Hao},
  journal={IEEE Transactions on Medical Imaging},
  volume={44},
  number={2},
  pages={656--667},
  year={2024},
  publisher={IEEE}
}

@inproceedings{zhou2025robust,
  title={Robust multimodal survival prediction with conditional latent differentiation variational autoencoder},
  author={Zhou, Junjie and Tang, Jiao and Zuo, Yingli and Wan, Peng and Zhang, Daoqiang and Shao, Wei},
  booktitle={Proceedings of the IEEE/CVF Conference on Computer Vision and Pattern Recognition},
  pages={10384--10393},
  year={2025}
}

@article{klambauer2017self,
  title={Self-normalizing neural networks},
  author={Klambauer, G{\"u}nter and Unterthiner, Thomas and Mayr, Andreas and Hochreiter, Sepp},
  journal={Advances in neural information processing systems},
  volume={30},
  year={2017}
}

@article{cui2026mgcm,
  title={MGCM: Multi-modal graph convolutional mamba for cancer survival prediction},
  author={Cui, Jiaqi and Li, Yilun and Shen, Dinggang and Wang, Yan},
  journal={Pattern Recognition},
  volume={169},
  pages={111991},
  year={2026},
  publisher={Elsevier}
}

@article{zhong2026diffusion,
  title={Diffusion-based cross-staining feature transformation for whole slide image analysis: From H\&E to IHC representation learning},
  author={Zhong, Jialong and Zhang, Miao and Liu, Leiye and Liu, Tingwei and Jiang, Jiahong and Piao, Yongri and Xu, Rui and Tian, Feng and Sun, Weibing and Bi, Huan and others},
  journal={Medical Image Analysis},
  pages={104138},
  year={2026},
  publisher={Elsevier}
}

@inproceedings{zheng2025gmmamba,
  title={GMMamba: Group Masking Mamba for Whole Slide Image Classification},
  author={Zheng, Tingting and Yao, Hongxun and Jiang, Kui and Xiao, Yi and Zhao, Sicheng},
  booktitle={Proceedings of the IEEE/CVF International Conference on Computer Vision},
  pages={9935--9944},
  year={2025}
}

@inproceedings{zheng2025m3amba,
  title={M3amba: Memory mamba is all you need for whole slide image classification},
  author={Zheng, Tingting and Jiang, Kui and Xiao, Yi and Zhao, Sicheng and Yao, Hongxun},
  booktitle={Proceedings of the IEEE/CVF Conference on Computer Vision and Pattern Recognition},
  pages={15601--15610},
  year={2025}
}

@article{zheng2025graphmamba,
  title={GraphMamba: Whole slide image classification meets graph-driven selective state space model},
  author={Zheng, Tingting and Yao, Hongxun and Zhao, Sicheng and Jiang, Kui and Xiao, Yi},
  journal={Pattern Recognition},
  volume={167},
  pages={111768},
  year={2025},
  publisher={Elsevier}
}

\end{document}